\documentclass{article}

\usepackage{arxiv}

\usepackage[utf8]{inputenc} % allow utf-8 input
\usepackage[T1]{fontenc}    % use 8-bit T1 fonts
\usepackage{hyperref}       % hyperlinks
\usepackage{url}            % simple URL typesetting
\usepackage{booktabs}       % professional-quality tables
\usepackage{amsfonts}       % blackboard math symbols
\usepackage{nicefrac}       % compact symbols for 1/2, etc.
\usepackage{microtype}      % microtypography
\usepackage{lipsum}
\usepackage{graphicx}
\graphicspath{ {./images/} }
\usepackage{float}
\usepackage[numbers]{natbib}
\usepackage[T1]{fontenc}
\usepackage{times}
\usepackage{booktabs} 
\usepackage{graphicx}
\usepackage{natbib}
\usepackage{graphics}
\usepackage{multirow}
\usepackage{array}
\usepackage{arydshln} 
\usepackage{hhline}   

\title{Evaluating Embeddings for One-shot Classification of Doctor-AI Consultations}

\author{
Olumide E. Ojo \\
Instituto Politécnico Nacional (IPN), \\
Centro de Investigación en Computación (CIC), \\
Mexico City, Mexico.\\
\texttt{olumideoea@gmail.com} \\
   \And
Olaronke O. Adebanji\\
Instituto Politécnico Nacional (IPN), \\
Centro de Investigación en Computación (CIC), \\
Mexico City, Mexico.\\
\texttt{olaronke.oluwayemisi@gmail.com} \\
\And
Alexander Gelbukh \\
Instituto Politécnico Nacional (IPN), \\
Centro de Investigación en Computación (CIC), \\
Mexico City, Mexico.\\
\texttt{gelbukh@cic.ipn.mx} \\
   \And
Hiram Calvo\\
Instituto Politécnico Nacional (IPN), \\
Centro de Investigación en Computación (CIC), \\
Mexico City, Mexico.\\
\texttt{hcalvo@cic.ipn.mx} \\
  \And
Anna Feldman\\
Montclair State University, \\
Montclair, USA.\\
\texttt{feldmana@montclair.edu} \\
}

\begin{document}
\maketitle
\begin{abstract}
Effective communication between healthcare providers and patients is crucial to providing high-quality patient care. In this work, we investigate how Doctor-written and AI-generated texts in healthcare consultations can be classified using state-of-the-art embeddings and one-shot classification systems. By analyzing embeddings such as bag-of-words, character n-grams, Word2Vec, GloVe, fastText, and GPT2 embeddings, we examine how well our one-shot classification systems capture semantic information within medical consultations. Results show that the embeddings are capable of capturing semantic features from text in a reliable and adaptable manner. Overall, Word2Vec, GloVe and Character n-grams embeddings performed well, indicating their suitability for modeling targeted to this task. GPT2 embedding also shows notable performance, indicating its suitability for models tailored to this task as well. Our machine learning architectures significantly improved the quality of health conversations when training data are scarce, improving communication between patients and healthcare providers.
\end{abstract}


\section{Introduction}
NLP has seen remarkable advances in recent years due to the development of large language models and the focus on improving text representations. An area of interest within this field is text classification~\cite{tan2022machine, adebanji2022sequential, DBLP:conf/semweb/TaOACG21, ojo2023transformer, ojo2022automatic, ojo-etal-2023-legend, ojo2022language}, a task that has practical applications in several domains, including politics~\cite{rozado2023danger}, health~\cite{shoham2023cpllm, ojo2023medai, tonja2022cic}, economics~\cite{ojo2020sentiment, ojo2021performance},  etc. Without training data, developing representations that can capture the features of text poses different challenges. Sectors like healthcare, where information exchange is crucial, can benefit from the development and implementation of these representations, often called embeddings. In this study, embedding techniques within one-shot classification systems are evaluated for their effectiveness in recognising AI-generated text.

Doctor-AI consultations between patients and healthcare providers have emerged as a viable route, using AI to help Doctors diagnose, prescribe treatments, and provide medical advice~\cite{fraile2023collaboration, LUO2023107413, sun2022multi, geng2023patient, mao2023automatic, wornow2023shaky, bao2023disc, van2023medicospace}. Text data generated from these interactions must be classified for a variety of applications, including authorship identification, medical record administration, and evaluating the legitimacy of online medical advice. The impact of accurate text classification in healthcare extends well beyond the immediate consultation process. Users can make informed decisions about the medical advice they receive online with the help of a correct classification, which allows them to identify the source of information.

There has been a great deal of interest in the one-shot classification task~\cite{Carloni_2023_ICCV, elnaggar2023sleep, xie2023position, balasundaram2023foreground, shi2023one}, where a single data instance is presented to a model for a rapid decision, usually involving a yes or no response. Specifically, the one-shot classification of Doctor-AI consultations poses unique challenges and opportunities, because it requires the development of robust text representations capable of distinguishing between the linguistic patterns and characteristics of Doctor-written texts and those generated by AI models. Medical records, as well as AI-assisted medical advice, must be distinguished to ensure credibility and reliability. Embeddings, or dense vector representations of text, are a key component for classification models to be successful. We use embedding techniques to determine the accuracy and potency of our one-shot classification systems for Doctor-AI consultations.

We explore the importance of various embedding methods and their performance in the one-shot classification of Doctor-AI consultations, thus improving healthcare consultations and online advice. As part of this study, a detailed analysis of model performance is carried out using various embeddings, and the effectiveness and limitations of these embeddings are discussed. We employ the accuracy, precision, recall and F1 scores to evaluate these models adequately. Study findings provided us with a useful reference for future analyses and discussions, improving the quality and safety of clinical discourse through improved accuracy and reliability of health information. Exploring this multifaceted problem not only advances the state-of-the-art, but also contributes to the broader discussion about how human expertise meets AI capabilities.

The contributions of this paper are summarized as follows:

\begin{itemize}
  \item We advanced our understanding of how embeddings affect the classification of Doctor-AI consultations, with implications for the quality and reliability of online medical advice.
  \item Multiple one-shot classification models were developed, eliminating the need for extensive training data.
 \item In our analysis, we classified Doctor-AI consultations as binary and multiclass categories.
\end{itemize}

In the following sections, we discuss the following topics. Section 2 is a comprehensive review of the related literature, setting the foundation for our study. In Section 3, we delve into the details of the experimental setup and presents the findings derived from the analysis of the datasets. In Section 4, we discuss the implications of our findings. Finally, in Section 5, we give a summary of our contributions, acknowledge the limitations encountered during the research, and provide directions for future research.

\section{Literature Review}
Human or AI-authored texts are difficult to identify online and require a collaborative approach that involves professionals from various fields. Research in text classification of this type has focused predominantly on the use of NLP techniques, with machine learning playing a vital role in capturing the inherent features of the text. We highlight the gaps in the existing literature that this current research seeks to address by conducting a comprehensive evaluation of embedding techniques for the specific task of one-shot classification in healthcare consultations.

Within the field of radiology, it is evident that multilingual and diverse large language models (LLMs) are poised to play a pivotal role in shaping the future of global healthcare delivery. According to~\cite{liu2023evaluating}, there is potential for enhanced healthcare services using AI on a global scale. Their study provided a solid foundation for further exploratory research. There is great potential to expand the applications of LLM in various medical specialties and advance the development of multimodal LLM. The latter has the ability to handle complex and varied data types, enabling a more holistic understanding of patient health. However, as we navigate this evolving landscape of LLMs, it is of paramount importance to conscientiously consider their effective implementation and ethical deployment. 

According to the work of~\cite{elnaggar2023sleep}, the posture of sleep was associated with a variety of health conditions, including nocturnal cramps and more serious musculoskeletal problems. Traditionally, in-clinical sleep assessments have focused primarily on monitoring vital signs, such as brain waves. Although wearable sensors equipped with inertial measurement units have been used for the classification of sleep posture, previous studies have been limited in scope, typically addressing only a handful of common postures, which are insufficient for comprehensive clinical evaluations. In response to these limitations, the authors introduced a novel framework for the classification of sleep postures, focusing on a minimal set of joint angle measurements. This framework is validated through two distinct experimental pipelines: the first involves computer animation to generate synthetic postural data, and the second involves a pilot study of human participants that uses custom-made miniature wearable sensors. Medical experts could understand the characterization of body posture made possible by the one-shot learning framework. The proposed framework achieved impressive overall accuracy, reaching levels as high as 100\% on synthetic data and 92.7\% on real data, placing it on par with the state-of-the-art data-intensive algorithms available in the existing literature.

The authors in~\cite{Ge2022FewshotLF} recognized the significant potential of few-shot learning (FSL) methods, particularly in light of the limited availability of annotated text for various medical topics. Their study delved into the existing landscape of FSL methods within the realm of medical NLP spanning articles published between January 2016 and August 2021, drawing from reputable sources like PubMed/Medline, Embase, ACL Anthology, IEEE Xplore Digital Library, and pre-print repositories such as medRxiv through Google Scholar. Inclusion criteria were established to identify articles that encompass FSL and medical text, and the review systematically summarized these articles based on their data sources, objectives, training sets sizes, primary methodologies, and evaluation techniques. The results of this review revealed 31 studies that met the defined inclusion criteria, all of which were published after 2018, with a notable increase in publications since 2020. Among the predominant tasks addressed in these studies were concept extraction/named entity recognition and text classification. In particular, a substantial portion of the studies (68\%) resorted to the creation of synthetic few-shot scenarios, frequently employing the MIMIC-III dataset. The commonalities among the FSL methods used included integration with attention mechanisms, prototypical networks, and meta-learning. The authors emphasized the potential benefits of specialized biomedical FSL datasets to facilitate method development and comparative evaluations. 

In~\cite{Carloni_2023_ICCV}, the article introduced a novel approach for the automatic classification of medical images. This method incorporated a convolutional neural network backbone and a causality extractor module that captures the cause-effect relationships between the image features. The study assessed the approach in low-data scenarios, employing one-shot learning and meta-learning procedures. Binary and multiclass classification experiments are conducted on a prostate MRI dataset, and the causality-driven module's effectiveness is validated through an ablation study and qualitative assessments. The findings highlighted the significant role of causal relationships in improving the model's ability to make reliable and interpretable predictions, making it a promising technique for medical image classification tasks. The study focused on improving classification performance, particularly in scenarios with limited training data, such as those encountered in medical image analysis. The research explored causation within the context of image data, demonstrating performance improvements over baseline methods, with notable enhancements in multiclass and specific AUROC metrics.

\cite{Zhang2023.06.08.23291121} discussed the result of a study carried out by the DS4DH group on the MEDIQA-Chat tasks at ACL-ClinicalNLP 2023. Their research combined classical machine learning, specifically the Support Vector Machine (SVM), to classify medical dialogues, with the incorporation of one-shot prompts using GPT-3.5. The approach involved using dialogues and summaries from the same category as prompts to generate summaries for new dialogues. The results exceeded the average benchmark score, which established a robust reference for the evaluation of performance in the field. The study highlighted the effectiveness of combining traditional machine learning techniques such as SVM with advanced language models such as GPT-3.5 to summarize medical dialogue. This hybrid methodology holds promise for improving patient care documentation and supporting healthcare professionals in informed decision making. 

In scenarios where classification errors have significant consequences, such as early detection of threat events with limited data, ensuring the reliability of classification results becomes paramount. \cite{he2023clur} delved into the realm of Uncertainty Estimation for Few-Shot Text Classification (UEFTC), which was relatively unexplored. In UEFTC, their goal was to develop models that can predict the uncertainty associated with a classification outcome, indicating the likelihood that it is incorrect, particularly in cases with limited samples. Traditional uncertainty estimation models in text classification often fall short when applied to UEFTC. These models typically require many training samples, whereas the few-shot setting of UEFTC provides only a few or even just one support sample for each class in a given episode. To address this challenge, the paper introduced Contrastive Learning from Uncertainty Relations (CLUR) as a solution. CLUR was designed to be trained with only one support sample for each class, aided by pseudo-uncertainty scores. In particular, CLUR autonomously learned these scores using a novel approach to uncertainty relations.

The work of~\cite{liao2023mask} introduced MaskBERT, a framework designed to enhance the few-shot learning capabilities of BERT-based models. Unlike the existing few-shot learning strategies, MaskBERT focused on selectively applying masks to text input, filtering out irrelevant information, and guiding the model to concentrate on discriminative tokens. In addition, a contrastive learning loss function was employed to improve the separability of different categories and the compactness of the same category representations. Experimental results in benchmark datasets demonstrate the effectiveness of MaskBERT, showing superior performance compared to other NLP methods. Ablation studies further validate the effectiveness of its components.

Embeddings improved the detection of suicide risk and depression on social media platforms in a study by ~\cite{GHOSAL20231631}. Identifying the risk of suicide and depressive content shared online has significant social implications, particularly for the younger generation. The authors introduced a novel framework that leverages fastText embedding for contextual analysis, TF-IDF vectors to gauge term relevance, and the XGBoost machine learning classifier for precise classification of depression and content related to suicidal risk. Their innovative approach yields promising results, achieving an AUC of 0.78 and a weighted F1 score of 0.71 on a Reddit dataset. The study demonstrates robust performance compared to various embedding models and classifiers, underscoring the effectiveness of the embedding approach in addressing the classification problem.

Our exploration of embeddings in one-shot classification for Doctor-AI consultations is contextualized by identifying the significance of one-shot classification tasks, the relevance of embeddings, and the evolving field of AI consultations. Our contribution lies in the empirical evaluation of various embeddings, with a particular focus on one-shot classification learning. It is our goal to improve both the quality and efficiency of Doctor-AI interactions to benefit both patients and healthcare providers.

\section{Experiments}
\subsection{Approach}
We adopted a diverse set of embedding techniques, including bag of words (BoW) in conjunction with Term Frequency-Inverse Document Frequency (TF-IDF), character n-grams, Word2Vec, GloVe, fastText, and GPT2 embeddings. 
BoW provides a quantitative representation of word occurrences, while character n-grams capture sequential patterns. GloVe, fastText, and GPT2 embeddings offer semantic representations that preserve contextual relationships among words. These embedding techniques transform words into continuous vector spaces. GloVe embeddings leverage global statistical information to represent words based on their semantic similarity, fastText considers subword information for robust handling of diverse vocabulary, and GPT2 embeddings employ a transformer architecture to provide contextualized representations. By incorporating these advanced embedding techniques into our study on One-shot Classification models for Doctor-AI Consultations, we aim to enhance the models' ability to understand and classify medical conversations by leveraging the rich semantic and contextual information encoded in the embeddings. As a result of their ability to handle diverse data structures and complexity, the Logistic Regression Classifier (LRC), Random Forest Classifier (RFC), Support Vector Machine (SVM), Naive Bayes Classifier (NBC), Decision Tree Classifier (DTC), Gradient Boosting Classifier (GBC), and Multilayer Perceptron (MLP) were selected. Each model was trained on the feature-extracted representations of doctor-AI consultations and evaluated using standard metrics, including Accuracy, Precision, Recall, and F1 scores. The selection of multiple models allows us to assess their relative strengths and weaknesses, providing a comprehensive understanding of their performance across different feature sets and embeddings.
\subsection{Dataset}
The traditional machine learning algorithms were applied to the MEDIC dataset ~\cite{ojo2023medai} and evaluated across different embeddings: Bag of Words (BoW), character n grams, Word2Vec, GloVe, fastText and GPT2. The dataset includes 1880 distinct texts for the Doctor's response, ChatGPT's response, and the rephrased Doctor's response, resulting in a total of 5640 text inputs. The dataset was split into three separate files. The first file contained responses from both the Doctor and the ChatGPT model (referred to as DC). The second file included responses from the Doctor and rephrased versions of the Doctor's responses (referred to as DR). The third file combined responses from Doctor, ChatGPT, and rephrased Doctor's responses (referred to as DCR). This file was used for the multi-label classification task. We chose a single instance of each unique label present in the dataset for model training, and applied these trained models to test on the remaining data. The embedding performance evaluation was carried out on the various proposed models using metrics such as accuracy, precision, recall, and F1 scores.

\subsection{Experiment Results}
Table~\ref{tab:table_one} presents the results of the one-shot classification models applied to the DC dataset using different embeddings. The traditional machine learning models (and its associated embedding features) evaluated in the table include the logistic regression classifier (LRC), random forest classifier (RFC), support vector machine (SVM), Gaussian naive Bayes classifier (NBC), decision tree classifier (DTC), gradient boosting classifier (GBC), and the multilayer perceptron network (MLP). For each model, the table shows the accuracy, precision, recall, and F1 score values, which serve as metrics to evaluate the model's performance. The Bag of Words (BoW), character n-grams, Word2Vec, GloVe, fastText, and GPT2 embeddings were used to transform text data into numerical vectors, enabling machine learning algorithms to process and classify them effectively.

In the DC dataset, MLP and Word2Vec embeddings demonstrated an accuracy of 96.44\%, along with impressive recall, precision, and F1 scores. The combination demonstrated that Word2Vec embeddings can effectively capture semantic representations. NBC and SVM with character n-gram embeddings also showed excellent performances, achieving an accuracy of 88. 30\% and 88. 86\% respectively, facilitating robust performance in the classification task. Character n-grams embeddings with SVM and NBC demonstrated strong performances with accuracies of 88.30\% and 88.86\%, respectively, showing the importance of taking into account the synergy between models and embeddings, as well as the interaction between semantic representations and classification algorithms. LRC and GPT2 embeddings are less effective, giving 57.15\% accuracy. Considering models and embeddings as a whole, the overall performance trends highlight the importance of systematic evaluation and selection, as different combinations exhibit substantial differences in performance.

\begin{table*}
\centering
\caption{Performance measures of the models in the DC dataset using various embeddings}
\small 
\renewcommand{\arraystretch}{1.2} 
\label{tab:table_one}
\begin{tabular}{:l:c:cc:cc:cc:cc}
\hhline{|=|*{8}{=}}
Model & Features & Accuracy & Precision & Recall & F1 \\
\hhline{|=|*{8}{=}}
LRC & BoW + TF-IDF & 0.7577 & 0.7596 & 0.7577 & 0.7573 \\ \hdashline
        & Character n-grams & 0.8830 & 0.8830 & 0.8830 & 0.8830 \\ \hdashline
        & Word2Vec & 0.8912 & \textbf{0.9104} & 0.8913 & 0.8899 \\ \hdashline
        & GloVe & \textbf{0.8928} & 0.8930 & \textbf{0.8928} & \textbf{0.8928} \\ \hdashline
        & fastText & 0.7316 & 0.8243 & 0.7318 & 0.7110 \\ \hdashline
        & GPT2 Embedding & 0.5715 & 0.5719 & 0.5716 & 0.5710 \\
\hhline{|=|*{8}{=}}
RFC & BoW + TF-IDF & 0.5253 & 0.7325 & 0.5250 & 0.3887 \\ \hdashline
        & Character n-grams & 0.4907 & 0.4790 & 0.4905 & 0.4099 \\ \hdashline
        & Word2Vec & \textbf{0.8524} & \textbf{0.8852} & \textbf{0.8525} & \textbf{0.8492} \\ \hdashline
        & GloVe & 0.7545 & 0.7603 & 0.7546 & 0.7532 \\ \hdashline
        & fastText & 0.7043 & 0.8129 & 0.7044 & 0.6762 \\ \hdashline
        & GPT2 Embedding & 0.6753 & 0.7758 & 0.6751 & 0.6429 \\
\hhline{|=|*{8}{=}}
SVM & BoW + TF-IDF & 0.7577 & 0.7595 & 0.7577 & 0.7573 \\ \hdashline
        & Character n-grams & 0.8830 & 0.8830 & 0.8830 & 0.8830 \\ \hdashline
        & Word2Vec & 0.8886 & \textbf{0.9086} & 0.8886 & 0.8872 \\ \hdashline
        & GloVe & \textbf{0.8928} & 0.8930 & \textbf{0.8928} & \textbf{0.8928} \\ \hdashline
        & fastText & 0.7614 & 0.8371 & 0.7616 & 0.7473 \\ \hdashline
        & GPT2 Embedding & 0.5715 & 0.5719 & 0.5716 & 0.5710 \\ 
\hhline{|=|*{8}{=}}
NBC & BoW + TF-IDF & 0.7577 & 0.7596 & 0.7577 & 0.7573 \\ \hdashline
        & Character n-grams & 0.8830 & 0.8830 & 0.8830 & 0.8830 \\ \hdashline
        & Word2Vec & 0.8926 & \textbf{0.9113} & 0.8926 & 0.8913 \\ \hdashline
        & GloVe & \textbf{0.8928} & 0.8930 & \textbf{0.8928} & \textbf{0.8928} \\ \hdashline
        & fastText & 0.7239 & 0.8300 & 0.7241 & 0.7014\\ \hdashline
        & GPT2 Embedding & 0.5715 & 0.5719 & 0.5716 & 0.5710 \\
\hhline{|=|*{8}{=}}
DTC & BoW + TF-IDF & 0.5021 & 0.5767 & 0.5019 & 0.3413 \\ \hdashline
        & Character n-grams & \textbf{0.9122} & \textbf{0.9253} & \textbf{0.9122} & \textbf{0.9116} \\ \hdashline
        & Word2Vec & 0.6915 & 0.7266 & 0.6916 & 0.6791 \\ \hdashline
        & GloVe & 0.7569 & 0.7571 & 0.7569 & 0.7569 \\ \hdashline
        & fastText & 0.8864 & 0.8878 & 0.8864 & 0.8863 \\ \hdashline
        & GPT2 Embedding & 0.6059 & 0.6314 & 0.6057 & 0.5856 \\
\hhline{|=|*{8}{=}}
GBC & BoW + TF-IDF & 0.5005 & 0.7502 & 0.5003 & 0.3340 \\ \hdashline
        & Character n-grams & 0.5074 & 0.7197 & 0.5072 & 0.3501 \\ \hdashline
        & Word2Vec & \textbf{0.9064} & \textbf{0.9206} & \textbf{0.9064} & \textbf{0.9056} \\ \hdashline
        & GloVe & 0.7918 & 0.7931 & 0.7917 & 0.7915 \\ \hdashline
        & fastText & 0.5880 & 0.7697 & 0.5883 & 0.5049 \\ \hdashline
        & GPT2 Embedding & 0.5093 & 0.5284 & 0.5091 & 0.4087 \\
\hhline{|=|*{8}{=}}
MLP & BoW + TF-IDF & 0.9130 & 0.9135 & 0.9130 & 0.9130 \\ \hdashline
        & Character n-grams & 0.9130 & 0.9135 & 0.9130 & 0.9130 \\ \hdashline
        & Word2Vec & \textbf{0.9644} & \textbf{0.9662} & \textbf{0.9644} & \textbf{0.9643} \\ \hdashline
        & GloVe & 0.8535 & 0.8565 & 0.8535 & 0.8532 \\ \hdashline
        & fastText & 0.8644 & 0.8925 & 0.8644 & 0.8619 \\ \hdashline
        & GPT2 Embedding & 0.6016 & 0.7365 & 0.6014 & 0.5351 \\ 
\hhline{|=|*{8}{=}} 
\end{tabular}
\end{table*}

Table~\ref{tab:table_two} offers the performance metrics for the one-shot classification model applied to the DR dataset using different models and features. The accuracy scores range from approximately 0.45 to around 0.83, illustrating a wide range of classification correctness. 

Using GPT2 Embedding, the MLP model showed remarkable accuracy, precision, recall, and F1 score values of 0.8356, 0.8690, 0.8356, and 0.8318. It demonstrates the effectiveness of leveraging contextualized embeddings provided by GPT2 for capturing intricate patterns in the DR dataset, creating a better prediction. DTC is the least-performing model when combined with BoW. Multiple models in this experiment illustrate the robustness and suitability of the GPT2 embedding for the DR dataset. FastText and GloVe display mixed results, with different results for different models. Moreover, the results indicate that the choice of embedding and model architecture is critical, as certain models, like MLP, exhibit substantial improvements over others, demonstrating the need to select embeddings and models carefully for optimal performance in datasets like this. The diversity in the scores emphasizes that no single model-embedding combination stands out as universally superior, particularly with the AI model's rephrasing of the Doctor's text.

\begin{table*}
\centering
\caption{Performance measures of the models in the DR dataset using various embeddings}
\small 
\renewcommand{\arraystretch}{1.2} % Adjust line spacing
\label{tab:table_two}
\begin{tabular}{:l:c:cc:cc:cc:cc}
\hhline{|=|*{8}{=}}
Model & Features & Accuracy & Precision & Recall & F1 \\
\hhline{|=|*{8}{=}}
LRC & BoW + TF-IDF & 0.5271 & 0.5271 & 0.5271 & 0.5271 \\
\hdashline
        & Character n-grams & \textbf{0.6128} & 0.6579 & \textbf{0.6130} & \textbf{0.5833} \\
        \hdashline
        & Word2Vec & 0.5950 & \textbf{0.6916} & 0.5952 & 0.5368 \\
        \hdashline
        & GloVe & 0.4745 & 0.4675 & 0.4743 & 0.4452 \\
        \hdashline
        & fastText & 0.4654 & 0.4622 & 0.4653 & 0.4540 \\
        \hdashline
        & GPT2 Embedding & 0.5327 & 0.5816 & 0.5329 & 0.4510 \\
\hhline{|=|*{8}{=}}
RFC & BoW + TF-IDF & 0.5564 & 0.5693 & 0.5563 & 0.5345 \\
\hdashline
        & Character n-grams & 0.5649 & 0.5869 & 0.5648 & 0.5353 \\
        \hdashline
        & Word2Vec & 0.5415 & 0.5475 & 0.5416 & 0.5270 \\
        \hdashline
        & GloVe & 0.5104 & 0.5104 & 0.5104 & 0.5101 \\
        \hdashline
        & fastText & 0.5186 & 0.5199 & 0.5187 & 0.5112 \\
        \hdashline
        & GPT2 Embedding & \textbf{0.6985} & \textbf{0.7047} & \textbf{0.6986} & \textbf{0.6962} \\
\hhline{|=|*{8}{=}}
SVM & BoW + TF-IDF & 0.5357 & 0.5357 & 0.5356 & 0.5355 \\
\hdashline
        & Character n-grams & \textbf{0.6128} & \textbf{0.6579} & \textbf{0.6130} & \textbf{0.5833} \\
        \hdashline
        & Word2Vec & 0.4840 & 0.4753 & 0.4839 & 0.4348 \\
        \hdashline
        & GloVe & 0.4745 & 0.4674 & 0.4743 & 0.4451 \\
        \hdashline
        & fastText & 0.4665 & 0.4633 & 0.4664 & 0.4550 \\
        \hdashline
        & GPT2 Embedding & 0.4947 & 0.4932 & 0.4948 & 0.4622 \\
\hhline{|=|*{8}{=}}
NBC & BoW + TF-IDF & 0.5271 &  0.5271  &  0.5271 &  0.5271 \\
\hdashline
        & Character n-grams & \textbf{0.6128} & \textbf{0.6579} & \textbf{0.6130} & \textbf{0.5833} \\
        \hdashline
        & Word2Vec & 0.4891 & 0.4840 & 0.4889 & 0.4465 \\
        \hdashline
        & GloVe & 0.4745 & 0.4674 & 0.4743 & 0.4451 \\
        \hdashline
        & fastText & 0.4702 & 0.4631 & 0.4701 & 0.4438 \\
        \hdashline
        & GPT2 Embedding & 0.4947 & 0.4932 & 0.4948 & 0.4622 \\
\hhline{|=|*{8}{=}}
DTC & BoW + TF-IDF & 0.5005 & 0.7502 & 0.5003 & 0.3340 \\
\hdashline
        & Character n-grams & 0.5027 & 0.7506 & 0.5029 & 0.3397 \\
        \hdashline
        & Word2Vec & 0.4758 & 0.4741 & 0.4757 & 0.4672 \\
        \hdashline
        & GloVe & 0.6035 & 0.6063 & 0.6036 & 0.6009 \\
        \hdashline
        & fastText & 0.5538 & 0.5553 & 0.5538 & 0.5506 \\
        \hdashline
        & GPT2 Embedding & \textbf{0.7326} & \textbf{0.7536} & \textbf{0.7325} & \textbf{0.7269} \\
\hhline{|=|*{8}{=}}
GBC & BoW + TF-IDF & 0.5005 & \textbf{0.7502} & 0.5003 & 0.3340 \\
\hdashline
        & Character n-grams & 0.5428 & 0.5967 & 0.5426 & 0.4684 \\
        \hdashline
        & Word2Vec & 0.4854 & 0.4804 & 0.4852 & 0.4516 \\
        \hdashline
        & GloVe & 0.5072 & 0.5074 & 0.5071 & \textbf{0.5025} \\
        \hdashline
        & fastText & 0.4729 & 0.4671 & 0.4727 & 0.4492 \\
        \hdashline
        & GPT2 Embedding & \textbf{0.5546} & 0.6427 & \textbf{0.5548} & 0.4736 \\
\hhline{|=|*{8}{=}}
MLP & BoW + TF-IDF & 0.5442 & 0.5505 & 0.5443 & 0.5298 \\
\hdashline
        & Character n-grams & 0.5820 & 0.6319 & 0.5821 & 0.5385 \\
        \hdashline
        & Word2Vec & 0.4537 & 0.4325 & 0.4536 & 0.4073 \\
        \hdashline
        & GloVe & 0.4614 & 0.4546 & 0.4613 & 0.4408 \\
        \hdashline
        & fastText & 0.4478 & 0.4417 & 0.4478 & 0.4331 \\
        \hdashline
        & GPT2 Embedding & \textbf{0.8356} & \textbf{0.8690} & \textbf{0.8356} & \textbf{0.8318} \\
\hhline{|=|*{8}{=}} 
\end{tabular}
\end{table*}

Table ~\ref{tab:table_three} presents a detailed overview of the performance measures of the one-shot classification models applied to the DCR dataset using various embeddings. Accuracy scores range from approximately 0.28 to 0.66, indicating substantial variation in overall classification correctness between different combinations of model embeddings. Among the best-performing combinations, the MLP model achieved an accuracy of 0.6642 with BoW. DTC embeddings with fastText gave the worst results with an accuracy of 0.2872 and low precision, recall, and F1 scores. This highlights the model's sensitivity to both embeddings and models, therefore, it is essential to carefully select the model for optimal outcomes. For the specific characteristics of the dataset at hand, it is important to conduct thorough experimentation and fine-tuning to identify the best embeddings and models.

This result reflects the substantial variations in performance, highlighting that model-embedding combinations at the lower end of the range are less effective on this specific dataset, whereas those at the higher end excel in classification. The detailed scores demonstrate that no model embedding combination is universally superior, and the choice depends on the type of dataset.

\begin{table*}
\centering
\caption{Performance measures of the models in the DCR dataset using various embeddings}
\small
\renewcommand{\arraystretch}{1.2} % Adjust line spacing
\label{tab:table_three}
\begin{tabular}{:l:c:cc:cc:cc:cc}
\hhline{|=|*{8}{=}}
Model & Features & Accuracy & Precision & Recall & F1 \\
\hhline{|=|*{8}{=}}
LRC & BoW + TF-IDF & 0.3989 & 0.4807 & 0.3986 & 0.2957 \\
\hdashline
& Character n-grams & 0.4172 & 0.5989 & 0.4169 & 0.3251 \\
\hdashline
& Word2Vec & \textbf{0.6564} & \textbf{0.6758} & \textbf{0.6563} & \textbf{0.6604} \\
\hdashline
& GloVe & 0.4691 & 0.4402 & 0.4690 & 0.4224 \\
\hdashline
& fastText & 0.4551 & 0.5760 & 0.4551 & 0.4620 \\
\hdashline
& GPT2 Embedding & 0.3895 & 0.3557 & 0.3897 & 0.3700 \\
\hhline{|=|*{8}{=}}
RFC & BoW + TF-IDF & 0.3340 & 0.4446 & 0.3337 & 0.1679 \\
\hdashline
& Character n-grams & 0.3523 & 0.4741 & 0.3520 & 0.2163 \\
\hdashline
& Word2Vec & 0.4842 & 0.5947 & 0.4843 & 0.4984 \\
\hdashline
& GloVe & 0.4716 & 0.4425 & 0.4717 & 0.4456 \\
\hdashline
& fastText & 0.5394 & \textbf{0.6011} & 0.5394 & \textbf{0.5531} \\
\hdashline
& GPT2 Embedding & \textbf{0.5741} & 0.5784 & \textbf{0.5742} & 0.5152 \\
\hhline{|=|*{8}{=}}
SVM & BoW + TF-IDF & 0.4741 & 0.4824 & 0.4739 & 0.4163 \\
\hdashline
& Character n-grams & \textbf{0.5956} & \textbf{0.6282} & \textbf{0.5956} & \textbf{0.5628} \\
\hdashline
& Word2Vec & 0.5380  & 0.6137 & 0.5377 & 0.5244 \\
\hdashline
& GloVe & 0.4044 & 0.4225 & 0.4043 & 0.4031 \\
\hdashline
& fastText & 0.4409 & 0.5676 & 0.4410 & 0.4439 \\
\hdashline
& GPT2 Embedding & 0.3652 & 0.3938 & 0.3653 & 0.3756 \\
\hhline{|=|*{8}{=}}
NBC & BoW + TF-IDF & 0.4741 & 0.4824 & 0.4739 & 0.4163 \\
\hdashline
& Character n-grams & \textbf{0.5956} & \textbf{0.6282} & \textbf{0.5956} & \textbf{0.5628} \\
\hdashline
& Word2Vec & 0.5442 & 0.6111 & 0.5439 & 0.5251 \\
\hdashline
& GloVe & 0.4044 & 0.4225 & 0.4043 & 0.4031 \\
\hdashline
& fastText & 0.4058 & 0.5460 & 0.4059 & 0.3850 \\
\hdashline
& GPT2 Embedding & 0.3652 & 0.3938 & 0.3653 & 0.3756 \\
\hhline{|=|*{8}{=}}
DTC & BoW + TF-IDF & 0.3363 & 0.5788 & 0.3360 & 0.1728 \\
\hdashline
& Character n-grams & 0.5041 & \textbf{0.6103} & 0.5038 & 0.4161 \\
\hdashline
& Word2Vec & \textbf{0.5239} & 0.5341 & \textbf{0.5238} & \textbf{0.4854} \\
\hdashline
& GloVe & 0.4452 & 0.4614 & 0.4453 & 0.4450 \\
\hdashline
& fastText & 0.2872 & 0.2280 & 0.2873 & 0.2346 \\
\hdashline
& GPT2 Embedding & 0.4205 & 0.4503 & 0.4206 & 0.4271 \\
\hhline{|=|*{8}{=}}
GBC & BoW + TF-IDF & 0.3340 & 0.7779 & 0.3337 & 0.1675 \\
\hdashline
& Character n-grams & 0.3340 & \textbf{0.7779} & 0.3337 & 0.1675 \\
\hdashline
& Word2Vec & \textbf{0.5848} & 0.5920 & \textbf{0.5845} & \textbf{0.5601} \\
\hdashline
& GloVe & 0.4924 & 0.4794 & 0.4922 & 0.4601 \\
\hdashline
& fastText & 0.5440 & 0.5741 & 0.5439 & 0.5519 \\
\hdashline
& GPT2 Embedding & 0.4496 & 0.5225 & 0.4495 & 0.3741 \\
\hhline{|=|*{8}{=}}
MLP & BoW + TF-IDF & \textbf{0.6642} & \textbf{0.6820} & \textbf{0.6641} & \textbf{0.6687} \\
\hdashline
& Character n-grams & 0.5587 & 0.5988 & 0.5588 & 0.5278 \\
\hdashline
& Word2Vec & 0.5834 & 0.5871 & 0.5831 & 0.5652 \\
\hdashline
& GloVe & 0.5085 & 0.5038 & 0.5083 & 0.5025 \\
\hdashline
& fastText & 0.3794 & 0.4653 & 0.3794 & 0.3581 \\
\hdashline
& GPT2 Embedding & 0.4643 & 0.4960 & 0.4644 & 0.4409 \\
\hhline{|=|*{8}{=}} 
\end{tabular}
\end{table*}

\begin{figure*}
  \centering
  \includegraphics[width=\textwidth]{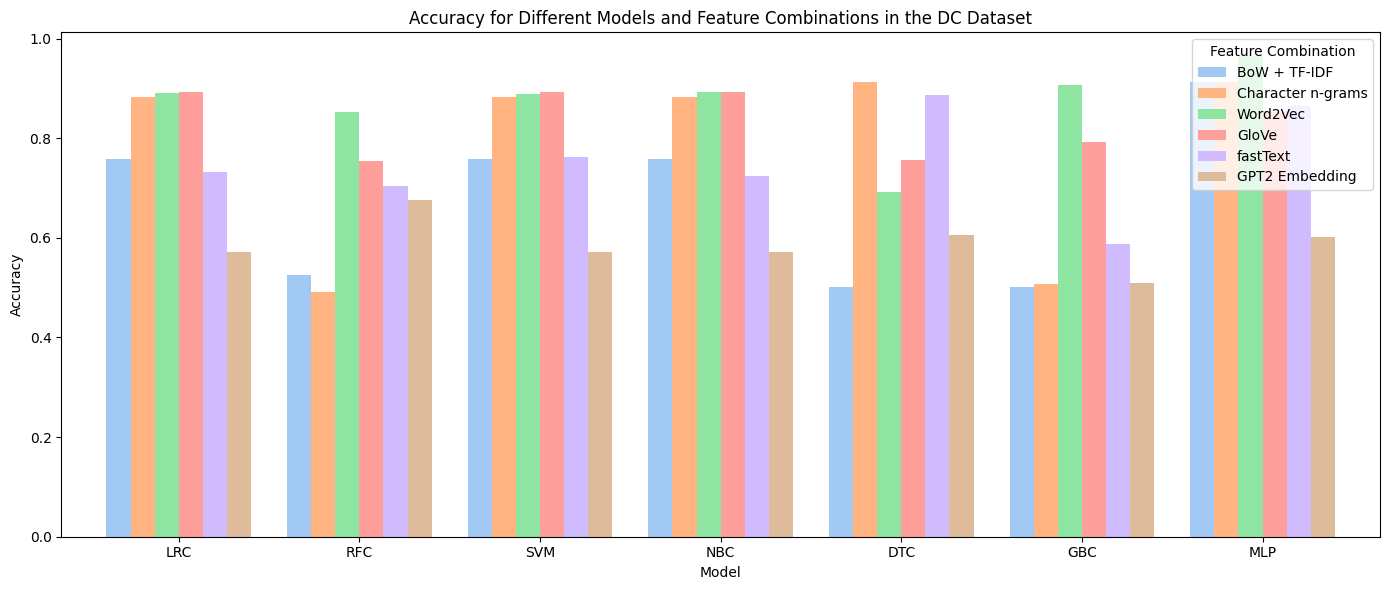} 
  \caption{Accuracy for Different Models and Feature Combinations in the DC Dataset}
  \label{fig:fig1}
\end{figure*}

\begin{figure*}
  \centering
  \includegraphics[width=\textwidth]{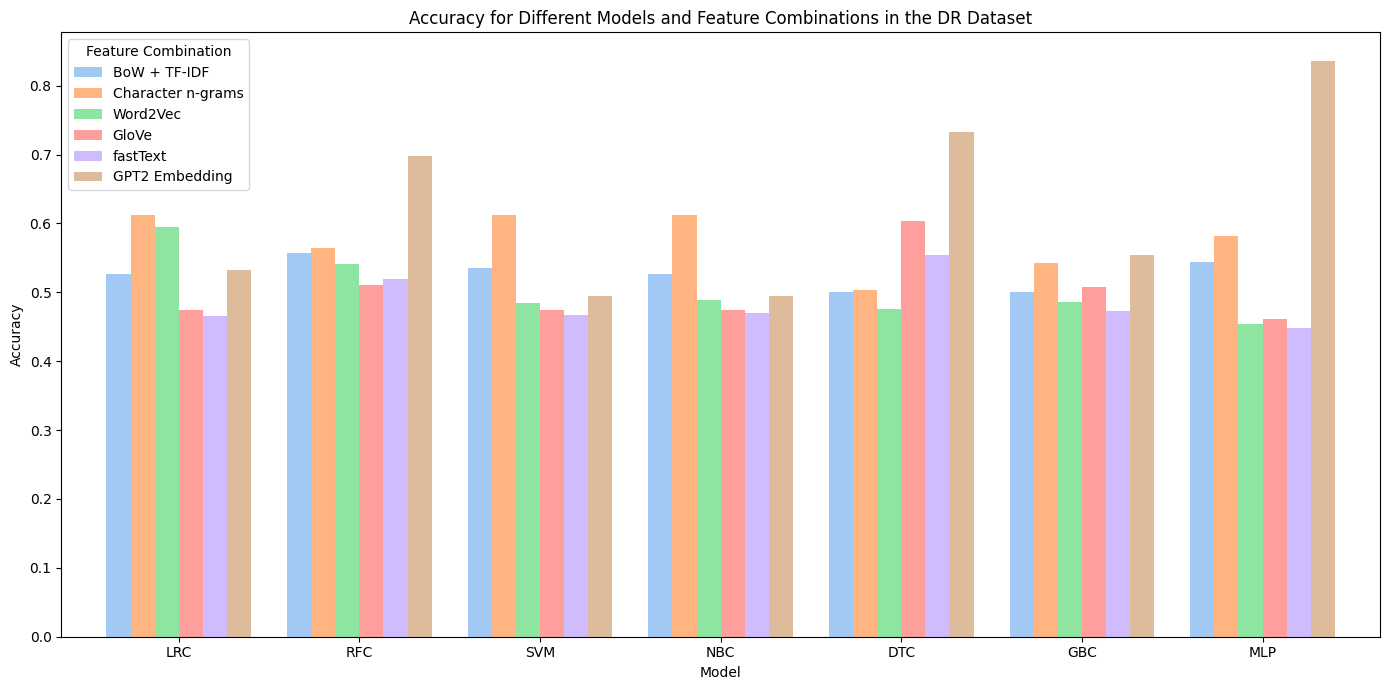} 
  \caption{Accuracy for Different Models and Feature Combinations in the DR Dataset}
  \label{fig:fig2}
\end{figure*}

\begin{figure*}
  \centering
  \includegraphics[width=\textwidth]{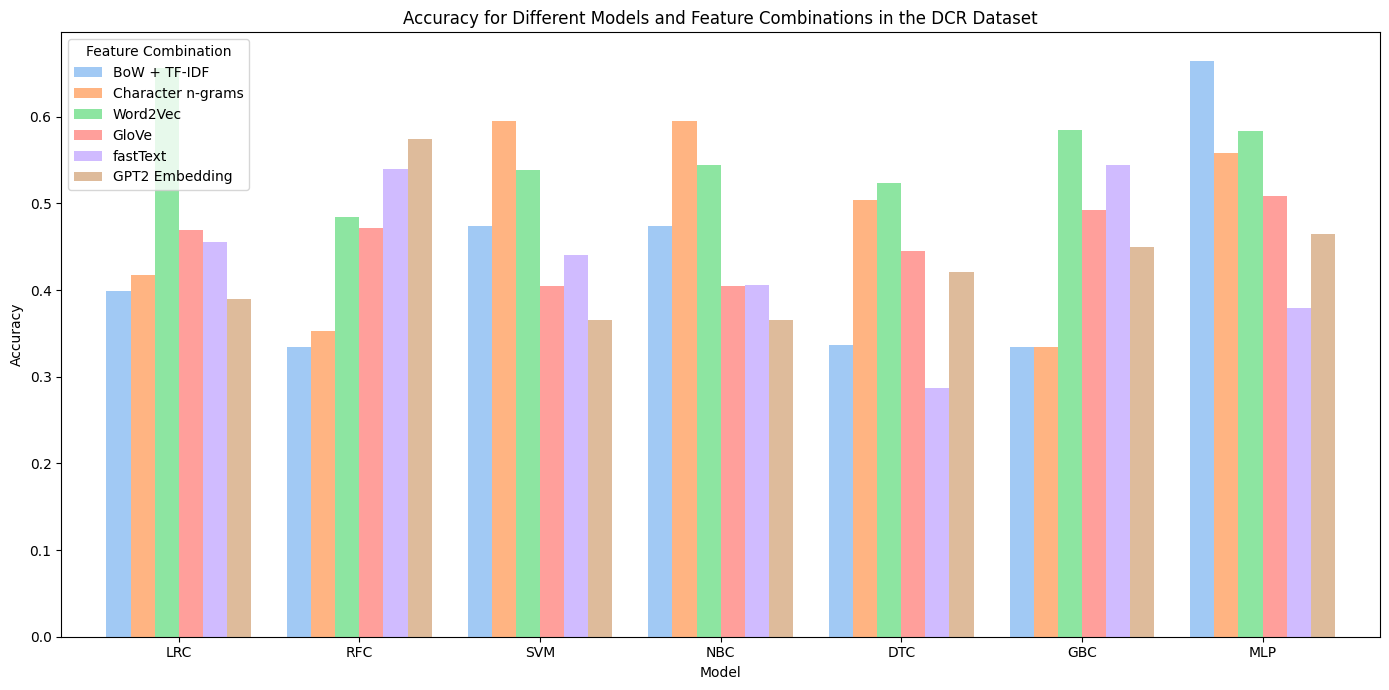} 
  \caption{Accuracy for Different Models and Feature Combinations in the DCR Dataset}
  \label{fig:fig3}
\end{figure*}

The bar graphs (Figures~\ref{fig:fig1}-~\ref{fig:fig3} visualize the accuracy scores for the seven machine learning models (LRC, RFC, SVM, NBC, DTC, GBC, MLP) applied using various feature combinations (BoW + TF-IDF, Character n-grams, Word2Vec, GloVe, fastText, GPT2 embedding). These visualizations aid in embedding selection and model decision, ultimately contributing to informed decisions and improved system decision for the datasets.

\begin{figure*}
  \centering
  \includegraphics[width=\textwidth]{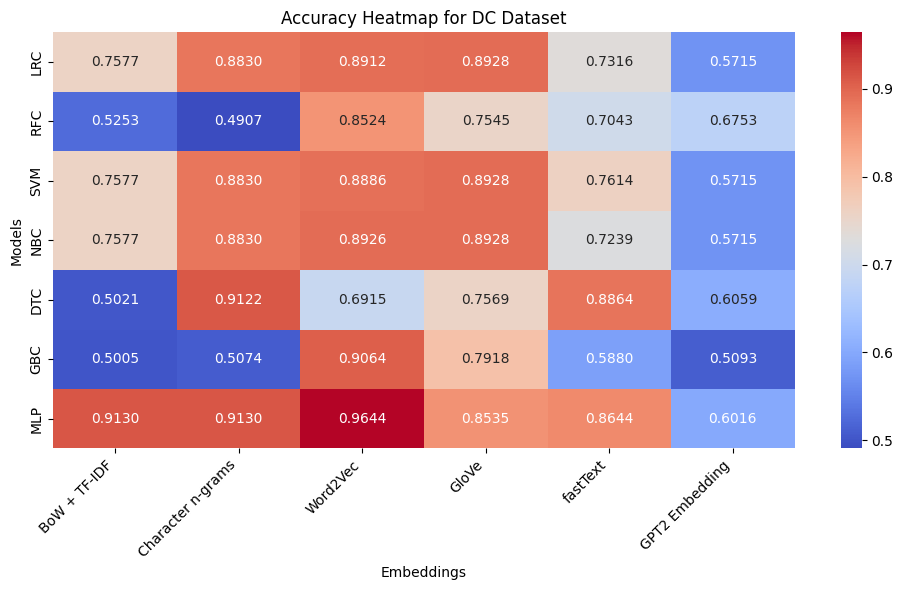} 
  \caption{Accuracy heatmap for DC Dataset}
  \label{fig:fig4}
\end{figure*}

\begin{figure*}
  \centering
  \includegraphics[width=\textwidth]{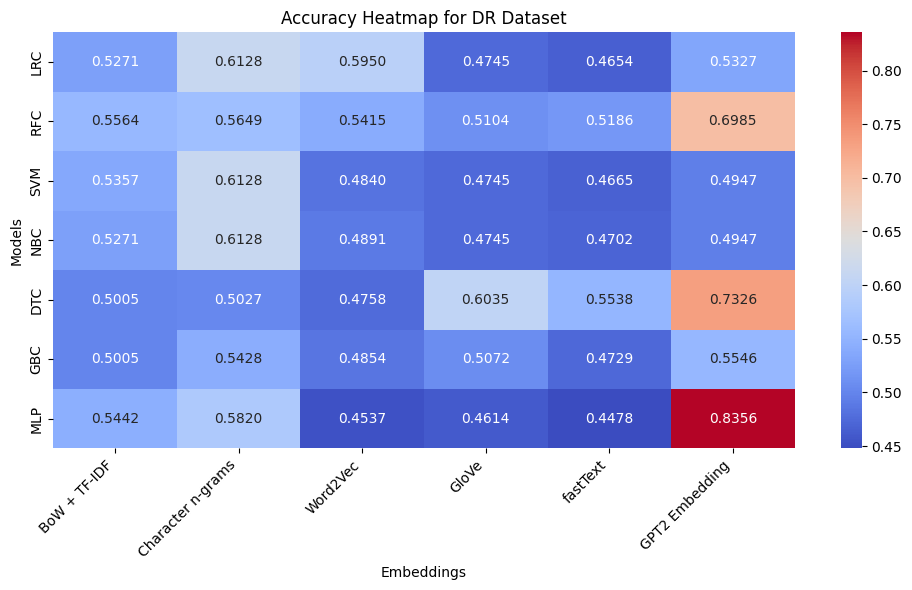} 
  \caption{Accuracy heatmap for DR Dataset}
  \label{fig:fig5}
\end{figure*}

For each dataset, we provide a visual summary of the performance of one-shot classification models across embeddings using Heatmaps. Fig. ~\ref{fig:fig4} represents the performance measures of the one-shot classification model for the DC dataset, showing the capabilities of the models across embeddings. This heatmap reveals which models and embeddings perform better and allows decisions to be made about the selection of models and embeddings for Doctor-AI consultations. Fig.~\ref{fig:fig5} illustrates how one-shot classification models perform on the DR dataset. With the graph, we can highlight trade-offs between models and embeddings within a particular context, and decide which models are best matched with which embeddings. Fig.~\ref{fig:fig5}, for the DCR dataset, sheds light on the performance of one-shot classification models and helps us to understand how different embeddings affect the performance of the model across various models using the multiclass dataset. By referencing the heatmaps, we show that certain embeddings consistently perform well on the chosen models.

\begin{figure*}
  \centering
  \includegraphics[width=\textwidth]{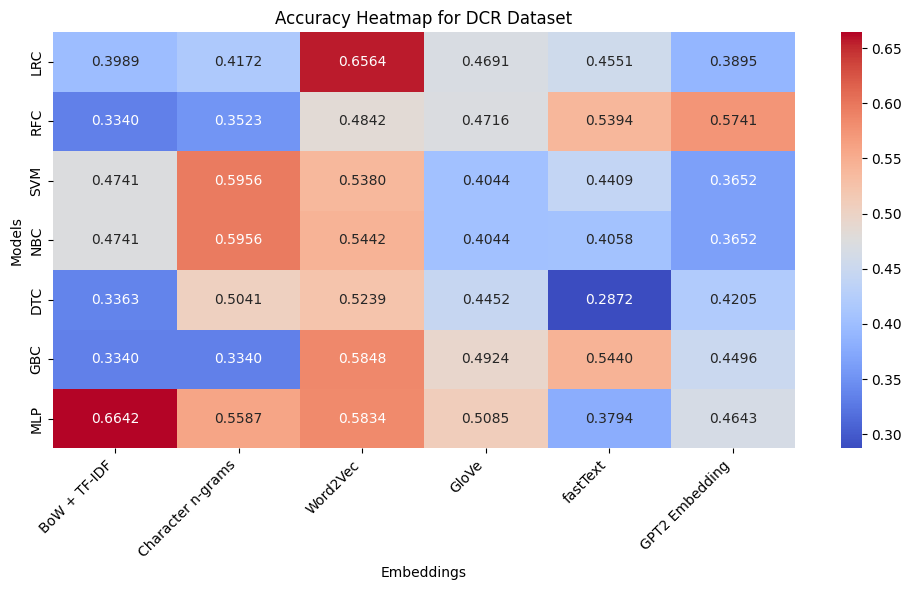} 
  \caption{Accuracy heatmap for DCR Dataset}
  \label{fig:fig6}
\end{figure*}

\section{Discussion}
We propose a comprehensive evaluation of embedding techniques for the one-shot classification of Doctor-AI consultations. The increasing adoption of AI in medical consultations has necessitated the development of efficient methods for distinguishing between responses generated by Doctors and those produced by AI systems. Different embedding techniques were tested and evaluated in our experiments to see how well they handle the unique challenges of one-shot learning in the classification of Doctor and AI-generated text.

In Table~\ref{tab:table_one}, which focuses on the DC dataset, Word2Vec and GloVe embeddings consistently exceed others in all models. This suggests their ability to achieve accurate classifications, underscoring their effectiveness in encoding semantic information from text data. Character n-grams also consistently deliver robust performance, serving as a dependable choice for encoding textual content. In contrast, the GPT2 embedding lags behind in some models, showing low accuracy and F1 scores. This discrepancy hints that it may not align effectively with the models designed for this specific one-shot classification task.

In Table~\ref{tab:table_two}, addressing the DR dataset, the accuracy ranges from approximately 0.45 to around 0.83, which indicates a substantial range in classification correctness. The precision metric varies significantly, highlighting the diversity in precision between classifications. These performance variations underscore the absence of a universally superior model-embedding combination. Instead, the results underscore the importance of selecting a combination of model embedding that aligns with the specific requirements of the one-shot classification task. Furthermore, the distinctive context of this dataset showcases the substantial influence of this context on the performance of the chosen models and embeddings.

Table~\ref{tab:table_three} provides an in-depth analysis of one-shot classification models applied to the DCR dataset using various embeddings. In this scenario, the accuracy scores range from approximately 0.28 to 0.66, highlighting substantial disparities in the overall classification accuracy for different combinations of modeling elements. Precision values illustrate a wide spectrum, varying from approximately 0.22 to 0.78, demonstrating the diversity in precision across classifications. Recall figures range from about 0.16 to 0.66, indicating different capabilities to capture true positives relative to actual positives. F1 scores, which balance precision and recall, range from roughly 0.17 to around 0.66, underlining the equilibrium between correct classification and the reduction of false positives and false negatives. The pronounced performance diversity in the DCR dataset emphasizes that there is no one-size-fits-all model-embedding combination. Instead, the choice of model and embedding should be customized to the specific dataset and task requirements. These results emphasize the critical role of context and data characteristics in determining the optimal model-embedding combination in one-shot classification scenarios.

\section{Conclusion}
This research has investigated the performance of various embedding methods in the context of one-shot classification for the analysis of Doctor-generated text and AI-generated text across three distinct datasets. We developed multiple one-shot classification models, which advanced our understanding of how embeddings impact the classification of text written by Doctors and text generated by AI, with implications for the quality and reliability of online medical advice.
We tested the following traditional machine learning models with different embedding techniques (bag-of-words, character n-grams, word2vec, GloVe, FastText, GPT2 embedding): logistic regression, naive bayes, random forest, support vector machines, decision tree, gradient boosting, multi-layer perceptron. The key findings are as follows:

\subsection{DC Dataset:}
\begin{itemize}
  \renewcommand{\labelitemi}{$\diamond$}
    \item Word2Vec and GloVe embeddings consistently outperformed other embeddings, achieving high accuracy, precision, recall, and F1 scores consistently exceeding 0.85.
    \item Character n-grams showed strong and consistent performance across certain models, making them a reliable alternative for capturing textual information.
    \item The GPT2 embedding consistently underperformed in a number of models, suggesting that it did not align well with the models developed and dataset of this kind.
\end{itemize}

\subsection{DR Dataset:}
\begin{itemize}
  \renewcommand{\labelitemi}{$\diamond$}
    \item The DR dataset exhibited a wide range of performance metrics across models and embeddings. This dataset demonstrates how effectively GPT2 embeddings capture intricate patterns in datasets of this kind, outperforming other embeddings across many models.
    \item Character n-grams contribute to higher accuracy, precision, recall, and F1 scores with certain models.
    \item SVM performance is consistent across different embeddings, suggesting its robustness across multiple feature representations, while Word2Vec maintains a competitive edge in accuracy across multiple models.
\end{itemize}

\subsection{DCR Dataset:}
\begin{itemize}
  \renewcommand{\labelitemi}{$\diamond$}
    \item Performance across the DCR dataset demonstrated substantial variations in overall classification correctness. GPT2 embedding faces challenges, often resulting in lower performance compared to other embeddings across multiple models.
    \item The choice of model-embedding combinations played a crucial role in classification effectiveness. Word2Vec consistently achieves competitive scores across a variety of models.
    \item These results underscore the importance of considering context and dataset characteristics when determining the optimal model-embedding combination. MLP consistently perform well in different embeddings, highlighting their stability in multiple feature representations.
\end{itemize}

\subsection{Implications:}
\begin{itemize}
  \renewcommand{\labelitemi}{$\diamond$}
  \item All embeddings perform well in all models, demonstrating their reliability and adaptability in capturing semantic features.
  \item The results emphasize the importance of selecting the appropriate embedding and model for machine learning depending on the dataset.
  \item The effectiveness of the models vary significantly with embedding choice, demanding careful consideration based on dataset features.
  \item Certain models may be more adaptable to specific types of feature representation because of their sensitivity to embeddings.
  \item Disparities in performance across dataset underscore application-specific considerations.
  \item What works well for one dataset may not be generalized to another, emphasizing the importance of evaluating embeddings in the context of the specific dataset and domain.
\end{itemize}

\subsection{Limitations:}
\begin{itemize}
  \renewcommand{\labelitemi}{$\diamond$}
  \item The present study is not without limitations, including potential biases in the dataset composition and inherent limitations in the choice of models and embeddings.
  \item Based on the data characteristics of the datasets used in this analysis (DC, DR, and DCR), generalizing these results to other datasets requires caution.
  \item The generalizability of the findings may be improved by more diverse and larger datasets.
  \item Language models such as GPT2 generate embeddings that are subject to updates and improvements, which can result in different performance results over time.
  \item The study does not address fine-tuned model architectures, which may further affect performance.
\end{itemize}

\subsection{Future Research Directions:}
In order to improve classification accuracy, we will explore advanced and context-aware embeddings in the future. We plan to explore the applicability of fine-tuned large-language model architectures, specifically tailored to the intricacies of health-related dialogues, to further improve classification. 

\section{Acknowledgments}
This work was done with partial support from the Mexican Government through the grant A1-S-47854 of CONACYT, Mexico, grants 20232138, 20230140, 20232080 and 20231567 of the Secretaría de Investigación y Posgrado of the Instituto Politécnico Nacional, Mexico. The authors thank CONACYT for the computing resources brought to them through the Plataforma de Aprendizaje Profundo para Tecnologías del Lenguaje of the Laboratorio de Supercómputo of the INAOE, Mexico and acknowledge the support of Microsoft through the Microsoft Latin America PhD Award.

\bibliographystyle{arxiv}
\bibliography{references}

\end{document}